\documentclass[letterpaper, conference]{ieeeconf}  

\IEEEoverridecommandlockouts                              
\usepackage[USenglish]{babel}
\usepackage{amsmath}
\usepackage{amssymb}
\usepackage{graphicx}

\def\R{\mathbb{R}} 
\def\N{\mathbb{N}} 
\def\Kinfty{\mathcal{K}_\infty} 
\def\K{\mathcal{K}}             

\def\Z{\mathbb{Z}} 
\def\O{\mathbb{O}} 
\def\B{\mathbb{B}} 
\def\path{p}       
\def\dimState{{n_x}} 
\def\dimInput{{n_u}} 
\def\dimPath{{n_p}}  
\def\uk{\mathbf{u}_k} 
\def\xk{\mathbf{x}_k} 
\def\l{\ell}         
\def\all{\underline{\alpha}_\ell} 
\def\alo{\underline{\alpha}_o}    
\def\hatalo{\hat{\underline{\alpha}}_o}    
\def\st{\textrm{s.t. }} 

\newtheorem{ass}{Assumption}
\newtheorem{lem}{Lemma}
\newtheorem{thm}{Theorem}

\newtheorem{rem}{Remark}

\usepackage{comment}
\usepackage{xcolor}

\title{MPC-based motion planning for non-holonomic systems\\ in non-convex domains}

\makeatother

\author{Matthias Lorenzen$^{1,\star}$, Teodoro Alamo$^2$, Martina Mammarella$^{3}$, Fabrizio Dabbene$^{3}$%
\thanks{$^{\star}${Corresponding author~{\tt\footnotesize matthias.lorenzen@hs-kempten.de}.}}%
\thanks{$^{1}$M. Lorenzen is with the University of Applied Sciences Kempten, Kempten, Germany.}%
\thanks{$^{2}$T. Alamo is with the Departamento de Ingenier\'ia de Sistemas y Autom\'atica, Universidad de Sevilla, Sevilla, Spain. 
}%
\thanks{$^{3}$F. Dabbene and M. Mammarella are with the CNR-IEIIT, Torino, Italy.}%
\thanks{T. Alamo acknowledges support from grant PID2022-142946NA-I00 and PID2022-141159OB-I00, funded by MICIU/AEI/ 10.13039/501100011033 and by ERDF/EU.}%
}

\begin{document}
\maketitle
\thispagestyle{empty}
\pagestyle{empty}

\begin{abstract}
  Motivated by the application of using model predictive control (MPC) for motion planning of autonomous mobile robots, a form of output tracking MPC for non-holonomic systems and with non-convex constraints is studied.
  Although the advantages of using MPC for motion planning have been demonstrated in several papers, in most of the available fundamental literature on output tracking MPC it is assumed, often implicitly, that the model is holonomic and generally the state or output constraints must be convex.
  Thus, in application-oriented publications, empirical results dominate and the topic of proving completeness, in particular under which assumptions the target is always reached, has received comparatively little attention.
  To address this gap, we present a novel MPC formulation that guarantees convergence to the desired target under realistic assumptions, which can be verified in relevant real-world scenarios.
\end{abstract}

\section{Introduction}
Model predictive control (MPC) has proven effective beyond classical process control tasks, showing, among other areas, particular promise for mobile robot motion planning in unstructured or dynamic environments.
A broad body of work has addressed path following, trajectory tracking or guaranteed obstacle avoidance.
Results have been derived for dynamic obstacles, e.g.,~\cite{Schoels2020_CIAO-MPCBasedSafeMotionPlanning, Soloperto2019_CollisionAvoidanceForUncertainNonlinearSysUsingMPC},
under uncertainty, e.g.,~\cite{Soloperto2019_CollisionAvoidanceForUncertainNonlinearSysUsingMPC, Gao2023_CollisionFreeMotionPlanningZeroOrderRobustOptimization},
and on numerically efficient re-formulations of the collision avoidance constraints either through approximations, e.g.,~\cite{Brito2019_MPCCForCollAvoidanceGuidance, Saccani2023_mtMPCForSafeUAVNavigation},
or exact formulations, e.g.,~\cite{Zhang2021_OptimizationBasedCollisionAvoidance, Helling2023_DualCollisionDetectionInMPCIncludingCullingTechniques}.

Nonetheless, deriving an MPC-based algorithm that guarantees target reaching under non-restrictive assumptions has received far less attention. Classical MPC convergence proofs typically rely on a terminal constraint, which would often result in a too long prediction horizon and therefore is frequently omitted in practice.
A widespread implementation can be found in the ROS2 software package NAV2~\cite{Macenski2020_Nav2}, which, e.g., implements as local controller the MPC algorithm presented in~\cite{Roesmann2015_TEBforMPC}.
However, without a suitable terminal constraint, most algorithms have a local and greedy nature, and thus can easily get trapped in local minima, as shown, e.g., in~\cite{Xiao2022_LearningMPC}.
%
%
Moreover, many general output tracking MPC formulations, e.g.,~\cite{Limon2018_NMPCForTracking}, crucially rely on convex constraints, which is generally violated in the presence of obstacles.
To circumvent this, attempts to handle obstacle avoidance using penalty methods have shown promising results in~\cite{Santos2023_NMPCForObstacleAvoidance}, but offer no strict guarantees on collision avoidance and target reaching.
In~\cite{Sanchez2021_NMPCPathFollowingWithObstAvoid}, the authors extend the results to solve the related problem of path following for constrained systems.
For exactly followable paths, convergence to the path and reaching of the goal is proven.
Furthermore, different simulation results suggest that the outcome suitably extends if the path is not exactly followable due to, e.g., obstacles or discontinuities.

%
The output-tracking MPC algorithm presented in~\cite{Soloperto2023_NonlinearMPCForOutputTrackingWithoutTerminalIngredients} can be used for robot navigation with guaranteed convergence to the target. However, the online optimization includes solving a global path planning problem, which significantly increases the complexity and is only tractable in very simple scenarios.
%

In contrast to the above cited papers, we present an algorithm which guarantees convergence to the target under hard state and input constraints, including obstacle avoidance, without the need to solve the global path planning problem at each iteration or assumptions which cannot be verified in the considered application. The main novelty and key concept is a different formulation of the optimization problem that includes an artificial reference tied to a reference path. While the results apply to general output tracking MPC, we focus on the application to motion planning of autonomous mobile robots.

The remainder of the paper is structured as follows.
Section~\ref{sec:probSetup} introduces the robot motion planning problem and definitions.
Section~\ref{sec:mainResults} presents the proposed MPC scheme alongside theoretical guarantees.
Simulation results in Section~\ref{sec:numEx} illustrate the main advantages, and concluding remarks are given in Section~\ref{sec:conclusions}.

\section{Problem Setup}\label{sec:probSetup}
The goal is to derive a control strategy to move an autonomous mobile robot (AMR) from an initial pose to a target pose while respecting kinematic and dynamic constraints and avoiding collisions with obstacles.
To solve this intricate task, a layered approach is usually taken, see, e.g.,~\cite{Macenski2020_Nav2}.
A global planner generates a reference path based on a map of the environment and a simplified robot model.
Several algorithms exist to this end, e.g., the well known A$^*$, D$^*$, or RRT algorithms (see~\cite{LaValle2006_PlanningAlgorithms} for an overview).
In a second layer, running at a higher frequency, this path is refined by a reactive local planner, based on the observed environment.
The final motor commands are then computed by low-level controllers.
In this paper, we address the middle layer, deriving a local planning and control algorithm.

The discrete-time, nonlinear dynamic system is given by 
\begin{equation}
  x_{k+1} = f(x_k,u_k),
  \label{eq:dynamics}
\end{equation}
where $f:\R^{\dimState}\times\R^{\dimInput} \mapsto \R^{\dimState}$ is a continuous function, modelling the dynamics of the robot and
$x_k \in \R^{\dimState}$ and $u_k \in \R^{\dimInput}$ are the state and input at sampling time k, respectively.
To model kinematic and dynamic constraints, e.g., maximum acceleration, velocity, or turn rate,
the state and input are assumed to be subject to constraints of the form
\begin{equation}
  (x_k, u_k) \in \Z,
  \label{eq:stateInputConstr}
\end{equation}
with a non-empty closed set $\Z \subset \R^{\dimState + \dimInput}$.
For technical reasons, it is assumed that the projection of $\Z$ onto the input space is compact.

To explicitly model obstacle avoidance,
we define the footprint of the robot, i.e., the space it occupies, at state $x$ by the set $\B(x) \subset \R^2$.
Moreover, we assume that the robot is operated in an environment with $N_o$ static obstacles, modelled by the sets $\O_i \subset \R^2$, $i=[1,N_o]$.
Collision avoidance constraints are then given by
\begin{equation}
  \B(x_k) \cap \O_i = \emptyset, \quad \forall i \in [1,N_o].
  \label{eq:collAvoidanceConstr}
\end{equation}


We assume that a global planner provides a reference path
\begin{equation}
  \path: [0,1] \mapsto \R^{\dimPath},
  \label{eq:globalRefPath}
\end{equation}
which is a continuous function, where $\path(0)$ is the initial configuration of the robot and $\path(1)$ is the target configuration.
We denote by $\mathcal{P}$ the image of $p$, i.e., all points of the path.
\begin{rem}
  Generally, global path planning algorithms do not produce optimal paths in finite time, especially when kinematic and dynamic constraints are considered. For this reason, we do not aim to track $\path$ exactly. Instead, it serves merely as guidance, ensuring that the MPC solution can be appropriately extended beyond the prediction horizon.
  %
\end{rem}

Next, we introduce the function $g:\R^\dimState \mapsto \R^\dimPath$, mapping the state of the robot to its configuration, and
function $g_p:[0,1] \mapsto \R^{\dimState}$, mapping a path variable $s\in[0,1]$ to the corresponding steady state.

\begin{ass}\label{ass:problemSetup}
  The following holds:
  \begin{enumerate}
    \item The function $f$ is continuous, $\Z$ is closed and the projection of $\Z$ onto the input space is compact.
    \item The function $p$ is continuous, the start of the path corresponds to the initial configuration of the robot, i.e., $\path(0)=g(x_0)$, and the end to the desired target configuration, i.e., $\path(1)=g(x_T)$.
    \item For each configuration $p(s)$ with $s \in [0,1]$, there exists a unique corresponding steady state $(x_s,u_s)$ in the interior of $\Z$ such that $g(x_s) = p(s)$ and $\B(x_s) \cap \O_i = \emptyset,~\forall i \in [1, N_o]$.
    \item\label{ass:itm:gp} There exists a Lipschitz continuous function $g_p: [0,1] \mapsto \R^{\dimState}$, such that $g \circ g_p = p$, with Lipschitz constant $L_{g_p}$.
  \end{enumerate}
\end{ass}
Note that the assumption related to the reference trajectory being in the interior of the feasible set is satisfied by many global path planning algorithms.
In case it is violated, 
the resulting path needs to be slightly perturbed or the planning must be performed considering enlarged objects. 

\section{Main Results}\label{sec:mainResults}
The main result of this paper is an MPC algorithm for which not only constraint satisfaction but also convergence to a desired target pose can be guaranteed.
The key concept is to rely on an artificial reference for the terminal constraint, inspired by~\cite{Limon2018_NMPCForTracking} and details from~\cite{Soloperto2023_NonlinearMPCForOutputTrackingWithoutTerminalIngredients}.
However, it is important to highlight that the results in~\cite{Limon2018_NMPCForTracking} crucially rely, among others, on two main assumptions: (i) convexity of the constraints, and (ii) either existence of a continuous terminal control law 
or controllability of the linearization. 

Since constraints~\eqref{eq:collAvoidanceConstr} generally lead to a non-convex constraint set, often assumption~(i)
does not hold in practical applications.
Furthermore, the dynamics of ground robots is often non-holonomic, which leads to a violation of assumption~(ii).
Hence, these results cannot be applied for motion control of an AMR with obstacle avoidance.

In order to circumvent these limitations, we present a novel MPC formulation. 
The major modification from state-of-the-art formulations is the usage of an artificial steady state, as in~\cite{Limon2018_NMPCForTracking}, which is tied to an artificial reference path that guarantees convergence to the target.

\subsection{MPC Algorithm}
In this framework, we define $\xk=(x_{0|k}, \ldots x_{N|k})$ and $\uk = (u_{0|k}, \ldots u_{N|k})$ as the predicted state and input sequences, respectively, which also represent the MPC optimization variables.
Moreover, we add another optimization variable $s_k \in [0,1]$, which defines the progress of the predicted terminal configuration at time $k$ along the path $\path$.
Furthermore, for readability and consistency with the existing literature, we add the optimization variable $x_{s_k} \in \R^{\dimState}$ and $u_{s_k} \in \R^{\dimInput}$ for the artificial steady state and input, which are constrained to be equal to $x_{N|k}$ and $u_{N|k}$, respectively.
However, to ease notation, we omit the dependency of $x_{s_k}$ and $u_{s_k}$ in the following cost function and MPC optimization.

For a given prediction horizon $N \in \N$ and state $x_k$, the MPC cost function is defined as
\begin{equation}
  J_N(\xk, \uk, s_k) =
  \sum_{l=0}^{N-1} \l(x_{l|k} -  x_{s_k}, u_{l|k} - u_{s_k}) + V_o(1-s_k),
  \label{eq:mpcCostFnc}
\end{equation}
with $x_{s_k} = x_{N|k}$, $u_{s_k} = u_{N|k}$, positive definite stage cost $\l:\R^{\dimState + \dimInput} \mapsto \R_{\ge 0}$ and offset cost $V_o: [0,1] \mapsto \R_{\ge 0}$.
The stage cost penalizes the tracking error of the predicted state and input with respect to $x_s$ and $u_s$, whereas the offset cost function is a measure of the distance to the target configuration along the reference path $\path$.
We introduce the following assumption for the stage and offset costs.
\begin{ass}\label{ass:cost}
  \leavevmode
  \begin{enumerate}
    \item The cost functions $\l$ and $V_o$ are continuous, $V_o$ is convex and $V_o(0)=0$.
    \item There exists a class $\Kinfty$ function $\all$ such that
          $\all(\|x\|) \le \l(x,u)$ for all $(x,u) \in \R^{\dimState + \dimInput}$.
    \item There exist a class $\K$ function $\alo$ such that
          $\alo(|{1-s}|) \le V_o(1-s)$ for all $s \in [0,1]$.
  \end{enumerate}
\end{ass}

At each time $k$, the proposed MPC scheme solves the following finite-horizon optimal control problem, i.e.,
%
\begin{subequations}
  \label{eq:mpcOpt}
  \begin{alignat}{2}
    \min_{\uk,\xk,s_k} & J_N(\uk, \xk, s_k)                                                  \\
    \st
                       & x_{0|k} = x_k                                                       \\
                       & x_{l+1|k} = f(x_{l|k}, u_{l|k}),      \quad l\in[0,N-1]             \\
                       & x_{s_k} = x_{N|k},~ u_{s_k} = u_{N|k}                               \\
                       & x_{s_k} = f(x_{s_k}, u_{s_k})                                       \\
                       & g(x_{s_k}) = \path(s_k)                                             \\
                       & (x_{l|k}, u_{l|k}) \in \Z,            \quad l\in[0,N]               \\
                       & \B(x_{l|k}) \cap \O_i = \emptyset,    \quad l\in[1,N],~i\in[1,N_o].
  \end{alignat}
\end{subequations}
Given the optimal solution $\uk^*, \xk^*, s_k^*$ to Problem~\eqref{eq:mpcOpt}, the MPC control law is defined as $\kappa_{MPC}(x_k) \doteq u_{0|k}^*$ and the optimal value function by $V_N(x_k) = J_N(\uk^*, \xk^*, s_k^*)$.
Similar to \cite{Limon2018_NMPCForTracking}, we employ the following controllability assumption, which asserts that from any feasible state in a neighborhood of a steady state $(x_s,u_s)$ on the path, the MPC cost is not excessive.
\begin{ass}\label{ass:controllability}
  There exist constants $b, \varepsilon > 0$ and $\sigma > 1$, such that for any $s\in[0,1]$ with corresponding steady state $x_s$, i.e., $x_s$ such that $g(x_s)=p(s)$, and for all $x_k \in \R^{\dimState}$ with $\|x_k-x_s\| \le \varepsilon$,
  there exists $(\uk,\xk)$ such that $(\uk,\xk,s)$ is feasible for Problem~\eqref{eq:mpcOpt} and
  $$\sum_{l=0}^{N-1} \l(x_{l|k} -  x_s, u_{l|k} - u_s) \le b\|x_k - x_s\|^{\sigma}.$$
\end{ass}
\medskip
We highlight that the kinematic constraints of many ground-based mobile robots are non-holonomic. Hence, the typical assumption of the linearized model being controllable is often not satisfied. However, our less strict Assumption~\ref{ass:controllability} can still be satisfied with a suitable stage cost $\ell$, as shown in Section~\ref{sec:numEx} for a differential wheeled robot.

The following theorem, which proof is provided in the appendix, summarizes the closed-loop system properties.
\begin{thm}\label{thm:mainResult}
  Suppose that for a given initial state $x_0$ and path $\path$, Assumptions~\ref{ass:problemSetup},~\ref{ass:cost}, and~\ref{ass:controllability} hold, and the system is controlled by the proposed MPC control law $\kappa_{MPC}$.
  If Problem~\eqref{eq:mpcOpt} is initially feasible, then it remains feasible for all $k>0$, the system satisfies \eqref{eq:stateInputConstr} and \eqref{eq:collAvoidanceConstr} throughout time,
  and the state converges to the target state $x_T = g_p(1)$.
\end{thm}

From a practical point of view, guaranteeing convergence to the target is a necessary condition to robustly apply the algorithm in real-world applications, where robots must not get stuck during operation.
To decrease the solution time of Problem~\eqref{eq:mpcOpt},
beyond carefully tuning the solver,
one could apply heuristics, e.g., to discard non-relevant obstacles as proposed in~\cite{Helling2023_DualCollisionDetectionInMPCIncludingCullingTechniques}.

We highlight that it is not strictly necessary for all obstacles to be known beforehand, provided that the path $\path$ can be re-computed at runtime, if needed.
Furthermore, in our simulations, often a solution could be found even if the global path went through obstacles, as long as the MPC prediction horizon was sufficiently long and similarly if dynamic obstacles - ignored by the path planner - were present.
This behavior is expected, as standard MPC formulations have generally been shown to perform well in such scenarios.
Nevertheless, we note that a rigorous extension of the algorithm, or of MPC in general, to handle these cases remains a relevant open question.

\section{Illustrative Example}\label{sec:numEx}
This section presents two numerical examples for a non-holonomic mobile robot, showing the advantage of the proposed MPC scheme and highlighting some important properties.
The first example focuses on avoiding local minima in an environment characterized by static obstacles, while the second demonstrates the difference to a path tracking formulation.


The robot is assumed to be a differential wheeled robot, which is common for industrial robots, e.g., MiR's autonomous mobile robots, as well as for simple educational robots like the Turtlebot.
The state $x = [p_x, p_y, \theta]^\top$ is the pose of the robot and the input $u = [v, \omega]$ consists of the linear velocity $v$ and the angular velocity $\omega$.
The movement is modelled by the differential equation
\begin{equation}
  \dot x =
  \begin{bmatrix} \dot{p}_x \\\dot{p}_y \\
    \dot{\theta}
  \end{bmatrix}=
  \begin{bmatrix}
    v \cos(\theta) \\
    v \sin(\theta) \\
    \omega
  \end{bmatrix},
  \label{eq:diffDriveDGL}
\end{equation}
where the mass of the robot is neglected, which is a reasonable assumption for small-scale robots or low velocities. As a result, the configuration space equals the state-space. For larger vehicles,
a dynamic model would be more suitable and could be used in exactly the same way with suitable mappings $g$ and $g_p$. Note that, for both models, there exists a steady state with input $u=0$ to each pose. To apply the proposed algorithm, the model~\eqref{eq:diffDriveDGL} is discretized using a 4th order Runge-Kutta integration method.
Moreover, the robot is subject to constraints on the input
\begin{equation*}
  \begin{bmatrix}
    -0.31 \\ -1.9
  \end{bmatrix}
  \le u \le
  \begin{bmatrix}
    0.31 \\ 1.9
  \end{bmatrix},
\end{equation*}
which are similar to the ones characterizing the Turtlebot 4.
We conclude that, Assumption~\ref{ass:problemSetup} is satisfied if the path is chosen appropriately, in particular not passing through any obstacles and starting at the initial pose.

In the following examples, we employ a cost function in the form of~\eqref{eq:mpcCostFnc} with stage cost
\begin{align*}
  \ell(x_{l|k} & -x_{s_k}, u_{l|k}-u_{s_k}) =                                          \\
               & (p_x - x_{s_k,1})^4 + (p_y - x_{s_k,2})^4 + 0.1(\theta - x_{s_k,3})^4 \\
               & + v^4 + \omega^4
\end{align*}
and offset cost
$V_o(1-s_k) = 1000(1-s_k)^2$.
The fourth power in the cost is chosen to satisfy Assumption~\ref{ass:controllability}.
It is well known, and can be easily checked, that the linearization of~\eqref{eq:diffDriveDGL} around any steady state is not controllable, although the nonlinear system is.
The system dynamics are input affine and the Lie bracket of the vector fields corresponding to the two inputs yields the necessary third vector field to span $\R^3$.
Practically speaking, the differential drive robot cannot move in lateral direction directly, but indirectly with, e.g., the typical `sideways parking maneuver'.
%
For an $\varepsilon$ small enough, this sideward movement with distance $\varepsilon$ can be generated applying the `Lie bracket input sequence', i.e.,
\begin{equation*}
  \uk = \left(
  \begin{bmatrix}0\\1\end{bmatrix},
  \begin{bmatrix}1\\0\end{bmatrix},
  \begin{bmatrix}0\\-1\end{bmatrix},
  \begin{bmatrix}-1\\0\end{bmatrix}
  \right)
\end{equation*}
with a magnitude of order $\sqrt{\varepsilon}$.
This leads to a stage cost of $\sum_{l=0}^{3} \ell(x_{l|k} - x_{s_k}, u_{l|k} - u_{s_k}) = k \varepsilon^2 + h.o.t.$.
Consequently, Assumption~\ref{ass:controllability} is satisfied with $\sigma=2$ if the path does not touch any obstacle.


\subsection{Obstacle Avoidance}
In the first example, a rectangular, axis-aligned box with corners $(1,-1)$, $(1.5,1)$ is placed as an obstacle between the start position $[0, 0]^\top$ and the target position $[2.5,0]^\top$.
The reference path is defined by
\begin{equation*}
  p(s) =
  \begin{bmatrix}
    2.5s,~ 1.1\sin(\pi s),~ \arctan(1.1\pi\cos(\pi s)/2.5)
  \end{bmatrix}^\top
\end{equation*}
so that it avoids the obstacle with some clearance from it.
The collision avoidance constraint~\eqref{eq:collAvoidanceConstr} is reformulated using Farkas' Lemma.
In particular, defining the obstacle as a set of linear inequalities $\O = \{p \in \R^2 ~|~ Ap \le b\}$, the constraint~\eqref{eq:collAvoidanceConstr} is equal to
\begin{equation}
  \label{eq:collAvoidanceConstrExample}
  \left(
  \begin{bmatrix}
    A & 0
  \end{bmatrix}
  x_{l|k} - b\right)^\top \mu_{l|k} \ge 0
\end{equation}
with the additional optimization variable $\mu_{l|k} \ge 0$ of appropriate size.

\begin{figure}
  \centering
  \includegraphics{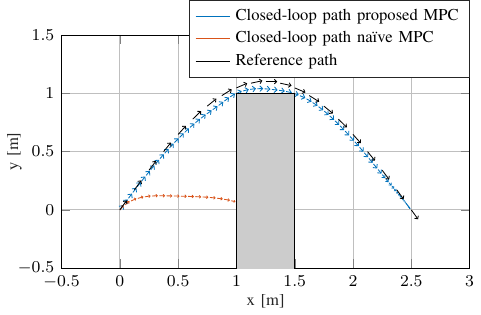}
  \caption{
    With the proposed implementation, the robot does move around the obstacle and reaches the target pose.
    In contrast, with an MPC implementation following~\cite{Limon2018_NMPCForTracking} and ignoring the assumption of a convex constraint set, the robot gets stuck in a local minimum in front of the target.
  }
  \label{fig:withobstacle}
\end{figure}

Figure~\ref{fig:withobstacle} shows the simulation result with a step size of $h=0.2s$ and prediction horizon of length $N=10$.
It can be observed that the robot avoids the obstacle and does reach the target, validating Theorem~\ref{thm:mainResult}.
The robot's dimensions are ignored here since for a circular footprint, like the one of the considered Turtlebot and many service robots, often the obstacles are simply inflated by the radius to consider a point-robot in collision avoidance.
More general footprints or tighter collision constraints can be taken into account using the methods summarized in, e.g.,~\cite{Helling2023_DualCollisionDetectionInMPCIncludingCullingTechniques}.
Furthermore, the collision avoidance constraint is only validated at each sampling time.

As highlighted in previous publications on using MPC algorithms for navigation and motion planning tasks, e.g.,~\cite{Brito2019_MPCCForCollAvoidanceGuidance,Xiao2022_LearningMPC}, the robot can easily get stuck in a local minimum when assumptions are violated.
Using the same setup as above and the tracking MPC algorithm presented in~\cite{Limon2018_NMPCForTracking} with only the target pose as tracking goal, the robot indeed gets stuck in front of the obstacle as shown in Figure~\ref{fig:withobstacle}.
Here, the assumption of a convex constraint set is violated, and thus the convergence results stated in the paper do not apply.
Although the initial orientation of the robot points towards avoiding the obstacle, it greedily moves towards the target position and finally remains in front of the obstacle.
While one could in principle apply a text-book MPC scheme with terminal equality constraint, the needed horizon length would be prohibitively large.
This shows the importance of a reference path or navigation function to avoid local minima instead of hoping for a global path planning solution using an MPC algorithm.

\subsection{Target reaching vs.\ path following}
In many mobile robot motion planning tasks, e.g., delivery of goods, close path tracking is not required but rather reaching the target while respecting the constraints.
However, tracking the path of the global planner is often implemented since this is a well studied topic.
The following example demonstrates that the proposed MPC-based motion planning algorithm does not simply track the reference path but that this is used only as a guidance to ensure that the target is reached.
The algorithm searches for an optimal solution, given the chosen cost.
Compared to the path $\path$, the robot can take shortcuts under the constraint that the prediction must end on the path.
Hence, the resulting degree of freedom strongly depends on the length of the MPC prediction horizon.
%

\begin{figure}
  \centering
  \includegraphics{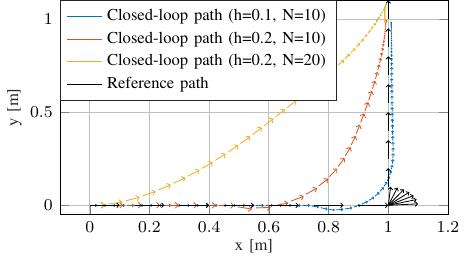}
  \caption{
    With a longer horizon in the optimal control problem, the robot can ``smooth out'' the reference path and finds a shorter route with a lower cost.
    The reference path is straight, $90^\circ$ turn, and straight again.
  }
  \label{fig:diffPredHorizon2}
\end{figure}

Figure~\ref{fig:diffPredHorizon2} shows the closed-loop trajectories that result from applying the proposed algorithm with different prediction horizons in terms of the number of prediction steps $N$ as well as different step sizes $h$.
It can be clearly seen that with a longer horizon in the optimal control problem~\eqref{eq:mpcOpt}, either in terms of larger $N$ or larger $h$, the robot finds a shorter path to the target, which has a lower cost, while increasingly deviating from the reference path.

\begin{figure}
  \centering
  \includegraphics{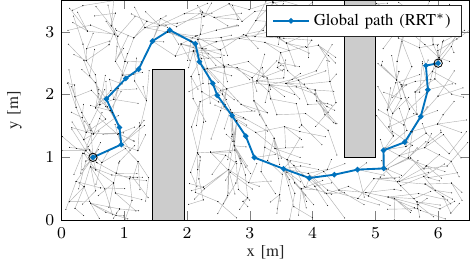}
  \caption{The result of a global path planner using an RRT$^*$ algorithm after 1000 iterations
    : It is generally characterized by many corners if a fast result is necessary, hence exact path tracking is often not desirable.}
  \label{fig:RRTexample}
\end{figure}

This property is of particular interest given that many global planning algorithms compute suboptimal solutions in finite time.
For example, popular RRT planning algorithms usually quickly find a global path which, however, is locally often very inefficient and characterized by unnecessary corners, see Figure~\ref{fig:RRTexample}.
In practice this path is often smoothed in a post processing step, a heuristic which requires some experience and testing.
With the proposed method, this smoothing is no longer required as demonstrated in Figure~\ref{fig:diffPredHorizon2}.


\section{Conclusions and Future Work}\label{sec:conclusions}
An MPC algorithm for motion planning and control of autonomous mobile robots has been presented.
With the proposed formulation of the online optimization problem, it can be guaranteed that the robot reaches the desired target pose and does not get stuck. 
This comes at the cost that a reference path needs to be available, which however, need not be optimal and is only employed for the artificial reference.


The algorithm presented is the foundation for future work.
An application of the proposed MPC-based navigation in a real-world application, where guaranteed, robust performance is necessary, is ongoing.
%
Furthermore, while we argue it is necessary, the strong dependence on the reference path is a limitation in dynamic environments or with unknown obstacles.
In these cases, it becomes desirable to allow further deviation from the reference path, while still guaranteeing convergence to the target.
This is an open question and important topic for future research.
It could probably be solved, if an answer is found on how obstacle size, length of the prediction horizon and existence of local minima are related.

\appendix
Before proving the main result of Theorem~\ref{thm:mainResult}, we first introduce and prove two necessary lemmas.

\begin{lem}\label{lem:convOfVnToVo}
  Suppose that Assumption~\ref{ass:controllability} is satisfied.
  If the sequence $(\|x_k - x_{s_k}^*\|)_{k>0}$ converges to 0, then $(V_N(x_k) - V_o(1-s_k^*)) \to 0$ for $k \to \infty$.
\end{lem}

\noindent
\textit{Proof:} Assume $\|x_k - x_{s_k}^*\| \to 0$, which implies that for any $\varepsilon > 0$ there exists $k \in \N$ such that $x_k$ is in an $\varepsilon$-neighborhood of $x_{s_k}^*$. Choose $\varepsilon$ such that Assumption~\ref{ass:controllability} can be applied.
Hence
\begin{equation*}
  \sum_{l=0}^{N-1} \l(x_{l|k}^* -  x_{s_k}^*, u_{l|k}^* - u_{s_k}^*) \le b\|x_k - x_{s_k}^*\|^\sigma
\end{equation*}
and thus $0 \le V_N(x_k) -  V_o(1-s_k^*) \le b\|x_k - x_{s_k}^*\|^\sigma \to 0$, which proves the claim.\hfill $\blacksquare$

\begin{lem}\label{lem:convergenceOfSkStar}
  Suppose that Assumption~\ref{ass:problemSetup},~\ref{ass:cost}, and~\ref{ass:controllability} are satisfied.
  If $s_k^*$ converges to some $s_\infty \in [0,1]$ and $(\|x_k - x_{s_k}^*\|)\to 0$ for $k\to\infty$,
  then $s_\infty = 1$.
\end{lem}
To prove the claim, we show that for $s_\infty \neq 1$, there exists an $s_k$ which is closer to 1 and has a lower overall cost.
As it will be shown, Assumption~\ref{ass:controllability} guarantees that the improvement in the offset cost $V_o$ is larger than the cost of moving from the state corresponding to  $s_\infty$ to the one corresponding to $s_k$.
In spirit, this is comparable to Assumption 4 in~\cite{Soloperto2023_NonlinearMPCForOutputTrackingWithoutTerminalIngredients}.
\noindent
\begin{proof}
  Define $\hat s_k = (1-\beta) s_k^* + \beta$ with $\beta \in [0,1]$.
  By Assumption~\ref{ass:controllability} and convergence of $(\|x_k - x_{s_k}^*\|)_{k\ge0}$,
  there exists $K\in\N$ and $\hat \beta \in (0,1]$
  such that for any $\beta \in [0, \hat \beta]$ and $k\ge K$
  there exists $(\uk,\xk)$ such that $(\uk,\xk,\hat s_k)$ is feasible for the MPC optimization~\eqref{eq:mpcOpt} and
  \begin{equation*}
    \sum_{l=0}^{N-1} \l(x_{l|k} -  x_{\hat s_k}, u_{l|k} - u_{\hat s_k}) \le b\|x_k - x_{\hat s_k}\|^\sigma.
  \end{equation*}
  With $\sigma > 1$, convexity implies
  \begin{align*}
    \|x_k - x_{\hat s_k}\|^{\sigma} & = \|(x_k - x_{s_k}^*) + (x_{s_k}^* - x_{\hat s_k})\|^\sigma                                  \\
                                    & = 2^\sigma \|\frac{1}{2}(x_k - x_{s_k}^*) + \frac{1}{2}(x_{s_k}^* - x_{\hat s_k})\|^\sigma   \\
                                    & \le 2^{\sigma-1}\|x_k - x_{s_k}^*\|^\sigma + 2^{\sigma-1}\|x_{s_k}^* - x_{\hat s_k}\|^\sigma
  \end{align*}
  and, from item~\ref{ass:itm:gp}) of Assumption~\ref{ass:problemSetup}, we infer
  \begin{equation*}
    \|x_{s_k}^* - x_{\hat s_k}\|^\sigma \le L_{g_p}^\sigma |s_k^* - \hat s_k|^\sigma
    = L_{g_p}^\sigma \beta^\sigma |s_k^* - 1|^\sigma.
  \end{equation*}
  Similarly, from convexity of $V_o$ and $V_o(0) = 0$, it follows
  \begin{align*}
    V_o(1-\hat s_k) & = V_o(1-((1-\beta) s_k^* + \beta))  \\
                    & = V_o(0\beta + (1-\beta) (1-s_k^*)) \\
                    & \le (1-\beta) V_o(1-s_k^*).
  \end{align*}
  Combining above inequalities leads to
  \begin{align}
    J(\uk,\xk,\hat s_k & ) \le 2^{\sigma-1}b\|x_k - x_{s_k}^*\|^\sigma \label{eq:upperBoundJ}                    \\
                       & + 2^{\sigma-1}bL_{g_p}^\sigma \beta^\sigma |s_k^* - 1|^\sigma + (1-\beta) V_o(1-s_k^*).
    \nonumber
  \end{align}
  Furthermore, the optimal value function can be bounded by
  \begin{equation}
    \begin{aligned}
      V_N(x_k) 
       & \ge \all(\|x_k - x_{s_k}^*\|) + V_o(1-s_k^*).
    \end{aligned}
    \label{eq:lowerBoundV}
  \end{equation}

  For notational convenience, we define the constants
  $c_{1,k} = 2^{\sigma-1}bL_{g_p}^\sigma |s_k^* - 1|^\sigma$,
  $c_{2,k} =  V_o({1-s_k^*})$, and
  the continuous function $e(\epsilon) = 2^{\sigma-1} b \epsilon^\sigma - \all(\epsilon)$.
  Combining~\eqref{eq:upperBoundJ} and~\eqref{eq:lowerBoundV} and using above definitions leads to
  \begin{multline}
    J(\uk,\xk,\hat s_k) - V_N(x_k) \le \\
    \underbrace{e(\|x_k - x_{s_k}^*\|)}_{\to 0} + \beta^\sigma \underbrace{c_{1,k}}_{\to c_1} - \beta \underbrace{c_{2,k}}_{\to c_2}.
    \label{eq:costDiff}
  \end{multline}
  By assumption $\|x_k - x_{s_k}^*\| \rightarrow 0$ and hence for $k \to \infty$, the first term converges to 0 and $c_{1,k}$ and $c_{2,k}$ to some value $c_1 \ge 0$ and $c_2 \ge 0$, respectively.

  For the sake of contradiction, assume $s_\infty \neq 1$ and hence $c_2 > 0$.
  Since $\sigma > 1$, this implies, that there exists $\beta \in (0, \hat \beta]$ such that $\beta^\sigma c_{1,k} < \beta c_{2,k}$ for $k$ large enough. Thus, with $\hat s_k$ corresponding to this $\beta$, the right hand side of~\eqref{eq:costDiff} is negative for $k$ large enough, which contradicts optimality of $V_N$.
\end{proof}

Next, we prove Theorem~\ref{thm:mainResult} relying on Lemma~\ref{lem:convOfVnToVo} and~\ref{lem:convergenceOfSkStar}.
\noindent
\begin{proof}
  Recursive feasibility and constraint satisfaction follow trivially by the usual arguments:
  A feasible solution to the optimization problem~\eqref{eq:mpcOpt} is given by the previously computed solution shifted by one time-step and appending the steady state and input at the end.
  The system constraint~\eqref{eq:stateInputConstr} is satisfied since $(x_k,u_k) = (x_{0|k},u_{0|k}) \in \Z$.
  Similarly, the obstacle avoidance constraints~\eqref{eq:collAvoidanceConstr} are satisfied since $x_{k+1} = x_{1|k}$ and hence $\B(x_{k+1}) \cap \O_i = \emptyset$ for $i=[1,N_o]$.

  To prove asymptotic stability, we first prove stability by showing that $V_N$ is a Lyapunov function, non-increasing along closed-loop trajectories. Then, we show that $x_k\rightarrow x_T$, making it an asymptotically stable equilibrium point. Note that $x_T$ is an equilibrium point of the closed-loop system.\\
  \noindent
  \textit{Lower bound on $V_N$:} We show the existence of a class $\Kinfty$ function that is a lower bound on the optimal value function $V_N$. In the following equation, the first inequality follows directly from Assumption~\ref{ass:cost}.
  To derive the second inequality, we employ the Lipschitz continuity condition from Assumption~\ref{ass:problemSetup}.
  For the third inequality, note that $\alo$ can be extended to a $\Kinfty$ function.
  Denoting this by $\hatalo$ we can define $\underline{\alpha}(s) = \min\{\all(s), \hatalo(L_{g_p}^{-1} s)\}$, which is of class $\Kinfty$, as well. Last, the fourth and fifth inequality follow from the weak triangle inequality for class $\K$ functions, see, e.g.,~\cite{Kellett2014_CompendiumOfComparisonFncResults}, and the standard triangle inequality, respectively.
  \begin{align*}
    V_N(x_k) & = \sum_{l=0}^{N-1} \l(x_{l|k}^* -  x_{s_k}^*, u_{l|k}^* - u_{s_k}^*) + V_o(1-s_k^*)                                                        \\
             & \ge \all(\| x_k -  x_{s_k}^* \|) + \alo(|1-s_k^*|)                                                                                         \\
             & \ge \all(\| x_k -  x_{s_k}^* \|) + \alo(L_{g_p}^{-1} \|x_T - x_{s_k}^*\|)                                                                  \\
             & \ge \underline{\alpha}(\| x_k -  x_{s_k}^* \|) + \underline{\alpha}(\|x_T - x_{s_k}^*\|)                                                   \\
             & \ge \frac{1}{2}\underline{\alpha}\big( \| x_k -  x_{s_k}^* \| + \|x_T - x_{s_k}^*\|\big) \ge \frac{1}{2}\underline{\alpha} \| x_k - x_T\|.
  \end{align*}
  \noindent
  \textit{Upper bound on $V_N$:}
  To derive an upper bound for $V_N$, we first derive a local upper bound.
  Let $x_k$ be in a neighborhood of $x_T$ where Assumption~\ref{ass:controllability} is satisfied and denote by $(\tilde{\mathbf{u}}_k, \tilde{\mathbf{x}}_k)$ the feasible input and state sequence that steers $x_k$ to $\tilde x_{N|k} = x_T$.
  Then, we have
  \begin{align*}
    V_N(x_k) & = \sum_{l=0}^{N-1} \l(x_{l|k}^* -  x_{s_k}^*, u_{l|k}^* - u_{s_k}^*) + V_o(1-s_k^*) \\
             & \le \sum_{l=0}^{N-1} \l(\tilde x_{l|k} -  x_T, \tilde u_{l|k} - u_T) + V_o(1-1)     \\
             & \le  b\|x_k - x_T\|^{\sigma}.
  \end{align*}
  This local upper bound can be extended to a class $\Kinfty$ function $\bar \alpha$ such that $V_N(x) \le \bar \alpha(\|x-x_T\|)$ for all feasible $x$ by standard arguments:
  Since $f$, $\l$, and $V_o$ are continuous, so is $J_N$.
  Hence, for any $x$ in a closed ball of radius $r$ around $x_T$, the maximum of $J_N$ over all admissible $u$ exists, is finite and monotonically increasing with $r$.
  An upper bound can thus be constructed as shown in the proof of~\cite[Prop. 5.7]{Gruene2017_NMPC}.\\
  \noindent
  \textit{Non-increase of $V_N$:}
  Let $\tilde{\mathbf{u}}_{k+1}, \tilde{\mathbf{x}}_{k+1}$ be the previously optimal solution $\uk^*, \xk^*$, shifted by one time step and $u_{N|k}^*$, $x_{N|k}^*$ appended at the end.
  Together with $\tilde s_{k+1} = s_k^*$, this is a feasible solution at time $k+1$ and
  \begin{align}
    \begin{split}
      V_N(x_{k+1}) - V_N(x_k) & \le J(\tilde{\mathbf{u}}_{k+1}, \tilde{\mathbf{x}}_{k+1}, s_k^*) - V_N(x_k) \\
                              & =                                                                           
      - \l(x_{0|k}-x_{s_k}^*,u_{0|k}-u_{s_k}^*)                                                             \\
                              & \le -\all(\|x_{0|k} - x_{s_k}^*\|) \le 0,
    \end{split}
    \label{eq:decrVN}
  \end{align}
  which implies stability.\\
  \noindent
  \textit{Convergence:} For asymptotic stability it remains to show that $x_k$ converges to $x_T$. Since $V_N$ is bounded from below, the sequence $(V_N(x_k))_{k\ge 0}$ converges to some limit $L$. Hence, the sequence $(\|x_k - x_{s_k}^*\|)_{k\ge 0}$ converges to 0 for $k\rightarrow\infty$. By Lemma~\ref{lem:convOfVnToVo} this implies $V_o(1-s_k^*) \to L$ and hence $s_k^*$ converges to some value $s_\infty \in [0,1]$. 
  By Lemma~\ref{lem:convergenceOfSkStar} $s_\infty = 1$ and hence $x_{s_k}^* \to x_T$ and thus $x_k \to x_T$.
\end{proof}

\bibliographystyle{IEEEtranNoUrl}
\bibliography{localBib}
\end{document}